\documentclass[conference]{IEEEtran}
\IEEEoverridecommandlockouts
\usepackage{cite}
\usepackage{amsmath,amssymb,amsfonts}
\usepackage{algorithmic}
\usepackage{graphicx}
\usepackage{textcomp}
\usepackage{xcolor}

\usepackage{amsmath, amsfonts, amssymb}
\usepackage{array}
\usepackage{booktabs}
\usepackage{enumitem}
\usepackage{flushend}
\usepackage{graphicx}
\usepackage{microtype}
\usepackage{multirow}
\usepackage{nicefrac}
\usepackage{printlen}
\usepackage{soul}
\usepackage{svg}
\usepackage{wrapfig}
\usepackage{lipsum}

\newcommand{\com}[1]{\iffalse #1 \fi}%

\newcommand{\noimage}{%
  \setlength{\fboxsep}{-\fboxrule}%
  \fbox{\phantom{\rule{100pt}{100pt}}File missing\phantom{\rule{100pt}{100pt}}}
}
\let\includegraphicsoriginal\includegraphics
\renewcommand{\includegraphics}[2][width=\textwidth]{\IfFileExists{#2}{\includegraphicsoriginal[#1]{#2}}{\noimage}}

\newcolumntype{+}{>{\global\let\currentrowstyle\relax}}
\newcolumntype{^}{>{\currentrowstyle}}

\emergencystretch=\maxdimen
\everypar{\looseness=-1 }

\newcounter{descriptcount}

\definecolor{PatternA}{RGB}{180, 22 , 0   }
\definecolor{PatternC}{RGB}{23 , 77 , 127 }
\definecolor{PatternB}{RGB}{55 , 144, 48  }

\usepackage{collcell}
\usepackage{xstring}
\usepackage{datatool}
\usepackage{booktabs}
\usepackage{environ}

\newcounter{CurrentRow}
\newcounter{CurrentColumn}
\newtoggle{DoneWithFirstRow}

\newcommand*{\FirstColumn}[1]{%
    \IfEq{\arabic{CurrentColumn}}{0}{%
        \global\togglefalse{DoneWithFirstRow}%
        \setcounter{CurrentRow}{1}
    }{%
        \global\toggletrue{DoneWithFirstRow}%
        \stepcounter{CurrentRow}%
    }%
    \setcounter{CurrentColumn}{0}%
    \NewData{#1}%
}
\newcommand*{\NewData}[1]{%
    \dtlexpandnewvalue%
    \stepcounter{CurrentColumn}%
    \iftoggle{DoneWithFirstRow}{%
        \dtlgetrow{TransposedTabularDB}{\arabic{CurrentColumn}}%
        \dtlappendentrytocurrentrow{\Alph{CurrentRow}}{#1}%
        \dtlrecombine%
    }{%
        \DTLnewrow{TransposedTabularDB}%
        \DTLnewdbentry{TransposedTabularDB}{\Alph{CurrentRow}}{#1}%
    }%
}%

\newcolumntype{F}{>{\collectcell\FirstColumn}c<{\endcollectcell}}
\newcolumntype{C}{>{\collectcell\NewData}{c}<{\endcollectcell}}

\newtoggle{EncounteredDataRow}

\newsavebox{\TempBox}
\DTLnewdb{TransposedTabularDB}

\NewEnviron{Ttabular}[1]{%
    \setcounter{CurrentColumn}{0}%
    \global\togglefalse{EncounteredDataRow}%
    \savebox{\TempBox}{%
        \begin{tabular}{FCCCCCC}
            \BODY%
        \end{tabular}%
    }%
    \begin{tabular}{#1}\toprule%
    \DTLforeach*{TransposedTabularDB}{\Aa=A, \Ba=B, \Ca=C}{%
      \DTLiffirstrow{}{\\\midrule}%
        \Aa & \Ba & \Ca %
    }\\\bottomrule%
    \end{tabular}%
}%

\newcolumntype{H}{>{\setbox0=\hbox\bgroup}c<{\egroup}@{}}

\usepackage{mathabx,graphicx}

\def\eqref#1{equation~\ref{#1}}

\def\1{\bm{1}}

\DeclareMathAlphabet{\mathsfit}{\encodingdefault}{\sfdefault}{m}{sl}
\SetMathAlphabet{\mathsfit}{bold}{\encodingdefault}{\sfdefault}{bx}{n}

\usepackage{pifont}
\usepackage{wasysym}

\usepackage{url}
\usepackage[T1]{fontenc}

\begin{document}

\title{A reading survey on adversarial machine learning: \\ Adversarial attacks and their understanding*\thanks{*Submitted as Term Project for Machine Learning Systems Engineering Course.}\\}

\author{\IEEEauthorblockN{Shashank Kotyan}
\IEEEauthorblockA{\textit{Laboratory of Intelligent Systems} \\
\textit{Department of Information Science and Engineerng}\\
Kyushu University, Fukuoka, Japan \\
Student ID: 2IE21054M; email: kotyan.shashank.651@s.kyushu-u.ac.jp}
}

\maketitle

\begin{abstract}
Deep Learning has empowered us to train neural networks for complex data with high performance.
However, with the growing research, several vulnerabilities in neural networks have been exposed.
A particular branch of research, Adversarial Machine Learning, exploits and understands some of the vulnerabilities that cause the neural networks to misclassify for near original input.
A class of algorithms called adversarial attacks is proposed to make the neural networks misclassify for various tasks in different domains.
With the extensive and growing research in adversarial attacks, it is crucial to understand the classification of adversarial attacks.
This will help us understand the vulnerabilities in a systematic order and help us to mitigate the effects of adversarial attacks.
This article provides a survey of existing adversarial attacks and their understanding based on different perspectives.
We also provide a brief overview of existing adversarial defences and their limitations in mitigating the effect of adversarial attacks.
Further, we conclude with a discussion on the future research directions in the field of adversarial machine learning.

\end{abstract}

~


\section{Introduction}

    \IEEEPARstart{N}{eural} networks have enabled us to obtain high performance in several applications across different domains like speech and text recognition (Natural Language Processing) and face, object and person recognition (Computer Vision).
    Most of these applications are only feasible with the assistance of neural networks either as a feature extractor for other machine learning methods or a classifier/detector on their own.

    However, despite their high performance against non-corrupted data, neural networks have been shown to misclassify critical adversarial samples in which perturbations (corruptions) are added to the original samples.
    Adversarial samples are noise-perturbed samples that can fail neural networks for tasks like image classification \cite{madry2018towards}, text classification \cite{tsai2019adversarial}, sound classification \cite{subramanian2019robustness}, and medical mole identifcation \cite{finlayson2019adversarial}.

    It was exhibited in \cite{szegedy2014intriguing} that neural networks behave oddly for almost the same images.
    Afterwards, in \cite{nguyen2015deep}, the authors demonstrated that neural networks show high confidence when presented with textures and random noise.
    This exposure led to discovering a series of vulnerabilities in neural networks, which were then exploited by using adversarial samples created by a particular class of algorithms known as adversarial attacks \cite{nguyen2015deep,brown2017adversarial,moosavi2017universal,su2019one}.

    \begin{figure*}[!t]
        \centering
        \includegraphics[width=\linewidth]{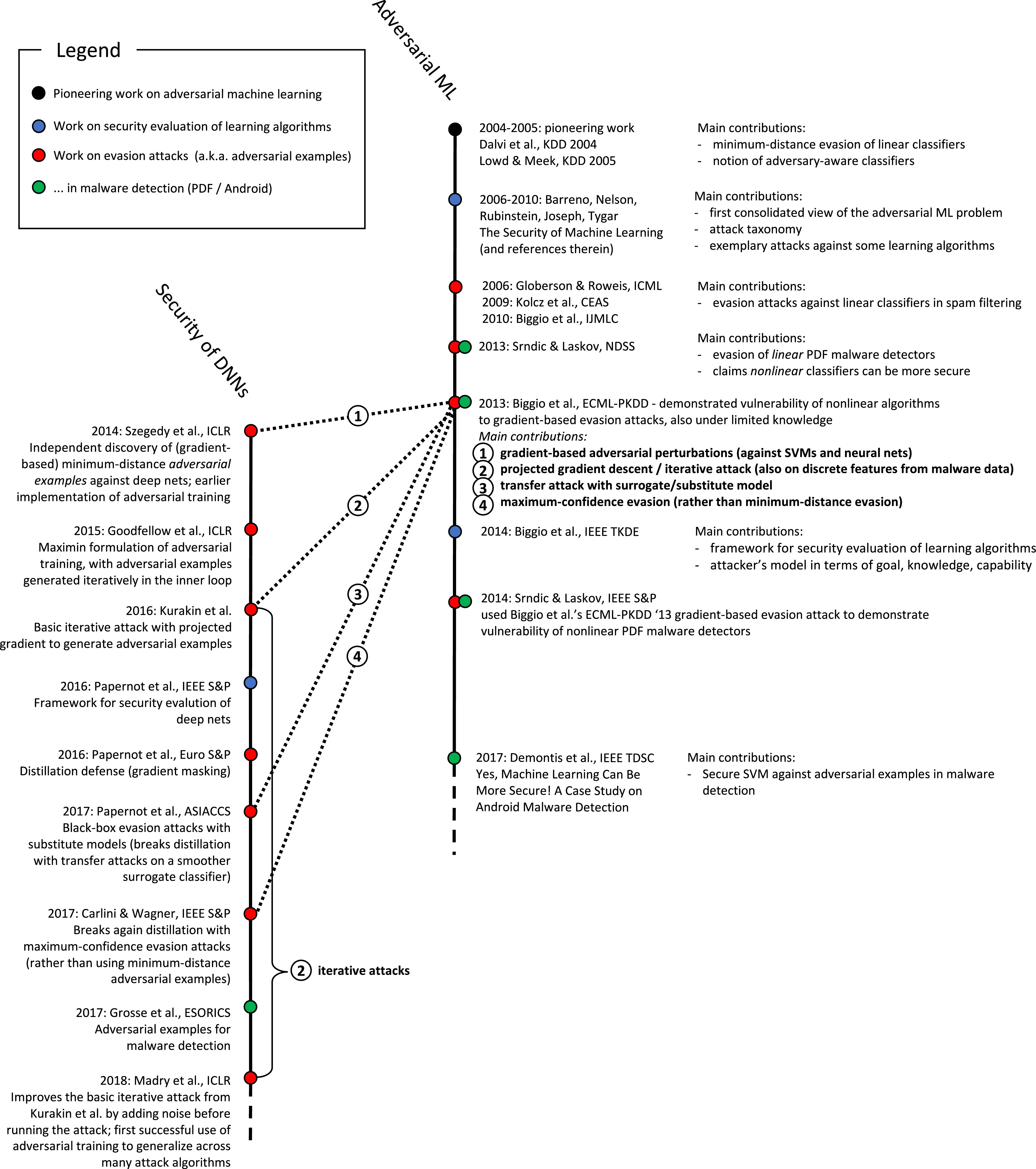}
        \caption{Timeline of adversarial attacks and samples in adversarial machine learning, compared to work on the security of deep networks \cite{biggio2018wild}. }
        \label{timeline}
    \end{figure*}

    From Universal perturbations, \cite{moosavi2017universal} that can be added to almost any image to generate an adversarial sample, to the addition of crafted patches \cite{brown2017adversarial} or in fact, even the addition of one-pixel \cite{su2019one} was also shown to cause networks to be enough to misclassify.
    Since they were discovered by \cite{szegedy2014intriguing} some years ago, both the quality and diversity of adversarial specimens have developed.
    A timeline of the development of adversarial attacks (and samples) is shown in Figure \ref{timeline}.

    These adversarial samples exhibit that conventional neural network architectures cannot understand concepts or high-level abstractions as we earlier speculated.
    Further, these adversarial attacks point out shortcomings in the reasoning of current machine learning algorithms.
    Thus, exposing our limited understanding of the working of a neural network.
    Also, we cannot explain the reason for predicting misclassified classes for the adversarial samples.

    Security and safety risks were also created by these adversarial attacks, which prohibits the use of neural networks.
    Most of these adversarial attacks can also be transformed into real-world attacks \cite{sharif2016accessorize,kurakin2016adversarial,athalye2017synthesizing,gu2017badnets}, which confer a big issue as well as a security risk for current neural networks' applications such as in autonomous vehicles.

    Many of these attacks can easily be made into real-world threats by printing out adversarial samples, as shown in \cite{kurakin2016adversarial}.
    Moreover, carefully crafted glasses can also be made into attacks \cite{sharif2016accessorize}.
    Alternatively, even general 3D adversarial objects were shown possible \cite{athalye2017synthesizing}.

    In fact, it was shown that by placing a few small stickers on the ground in an intersection, researchers showed that they could cause a self-driving car to make an abnormal judgment and move into the opposite lane of traffic \cite{yan2016can}.
    Also, placing a few pieces of tape can deceive a computer vision system into wrongly classifying a stop sign as a speed limit sign \cite{gu2017badnets}.

    Despite the existence of many variants of defences to these adversarial attacks \cite{goodfellow2014explaining,huang2015learning,papernot2016distillation,dziugaite2016study,hazan2016perturbations,das2017keeping,guo2018countering,song2018pixeldefend,xu2017feature,madry2018towards,ma2018characterizing,buckman2018thermometer}, no known learning algorithm or procedure can defend consistently \cite{carlini2017towards,tramer2018ensemble,athalye2018obfuscated,uesato2018adversarial,vargas2019robustness,tramer2020adaptive}.
    This shows that a more profound understanding of the adversarial attacks is needed to formulate consistent and robust defences.

    Several works have focused on understanding the reasoning behind such a lack of robust performance.
    It is hypothesised in \cite{goodfellow2014explaining} that neural networks' linearity is one of the main reasons for failure.
    Other investigation by \cite{thesing2019ai} shows that with deep learning, neural networks learn false structures that are simpler to learn rather than the ones expected.

    In \cite{sabour2015adversarial}, it is discussed that an adversarial sample may have a different interpretation of learned features than the benign sample.
    The authors show that learned features of adversarial samples are remarkably similar to different images of different true-class and links adversarial robustness to features learned by deep neural networks.
    Moreover, research by \cite{vargas2019understanding} unveil that adversarial attacks are altering where the algorithm is paying attention.

    The field of adversarial machine learning has contributed towards the development of some tools which could be helpful in the development of assessment for robustness.
    However, the sheer number of scenarios: attacking methods, defences, and metrics, make the current state-of-the-art difficult to perceive.
    It turns out that a simple robustness assessment is a daunting task, given the vast amount of possibilities and definitions along with their exceptions and trade-offs.

    However, most recent adversarial attacks and defences are white-box which can not be used to assess hybrids, non-standard neural networks and other classifiers in general and limit us in exploring the cases for our understanding.
    Improvements in robustness should also result in learning systems that can better reason over data and achieve a new level of abstraction.

    To better understand the adversarial attacks, here we provide a survey of adversarial attacks and their understanding.
    We classify the adversarial attacks based on the knowledge available to adversarial attacks, goals of adversarial attack, the scope of the adversarial attack, the strategy employed by the adversarial attack, optimisation used by adversarial attack, and constraints on perturbation imposed by adversarial attacks.

\section{Formal definition of adversarial samples}

    \begin{figure}[!t]
        \centering
        \includegraphics[width=\linewidth]{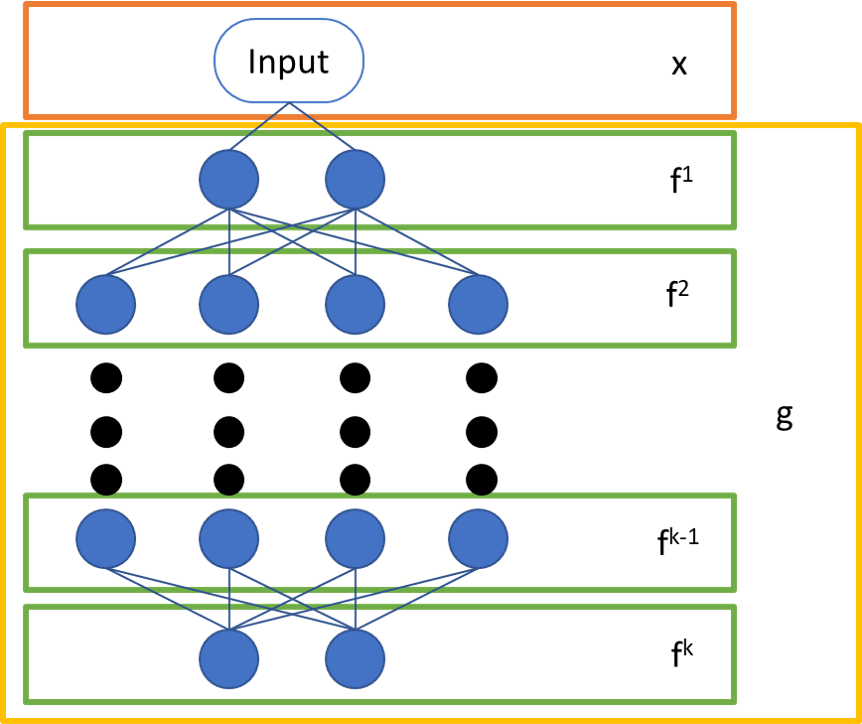}
        \caption{Visualisation of formal definition for Neural Network}
        \label{nn}
    \end{figure}

    Let us suppose that for the image classification problem, $x \in \mathbb{R}^{m \times n \times c}$ be the image\footnote{Here, we use the example image for $x$; however, the definition can be extended to other domains like text, and speech.} which is to be classified.
    Here $m, n$ is the image's width and height, and $c$ is the number of colour channels.

    A neural network comprises several neural layers composed of perceptrons (artificial neurons) linked together.
    Each of these perceptrons maps a set of inputs to output values with an activation function.

    Thus, function of the neural network (formed by a chain) can be defined as:
    \begin{equation}
    \begin{aligned}
    g(x) = f^{(k)}( \ldots f^{(2)}(f^{(1)}(x)))
    \end{aligned}
    \label{eqnn}
    \end{equation}
    where $f^{(i)}$ is the function of the $i^{\text{th}}$ layer of the network, where $ i = 1,2,3, \ldots , k$ and $k$ is the last layer of the neural network as shown in Figure \ref{nn}.
    In the image classification problem, $g(x) \in \mathbb{R}^{N}$ is the probabilities (confidence) for all the available $N$ classes.

    Also, in adversarial machine learning, adversarial samples $\hat{x}$ are defined as:
    \begin{equation}
    \begin{aligned}
    & \hat{x} = x + \epsilon_{x} \\
    & \{ \hat{x} \in \mathbb{R}^{m \times n \times 3} \mid {\operatorname{argmax}}[g(x)] \ne {\operatorname{argmax}}[g(\hat{x})]  \}
    \end{aligned}
    \label{adv}
    \end{equation}
    in which $\epsilon_{x}$ is the perturbation added to the input.

\section{Non-exhaustive classification of adversarial attacks}

\begin{figure*}[!t]
    \centering
    \includegraphics[width=\linewidth]{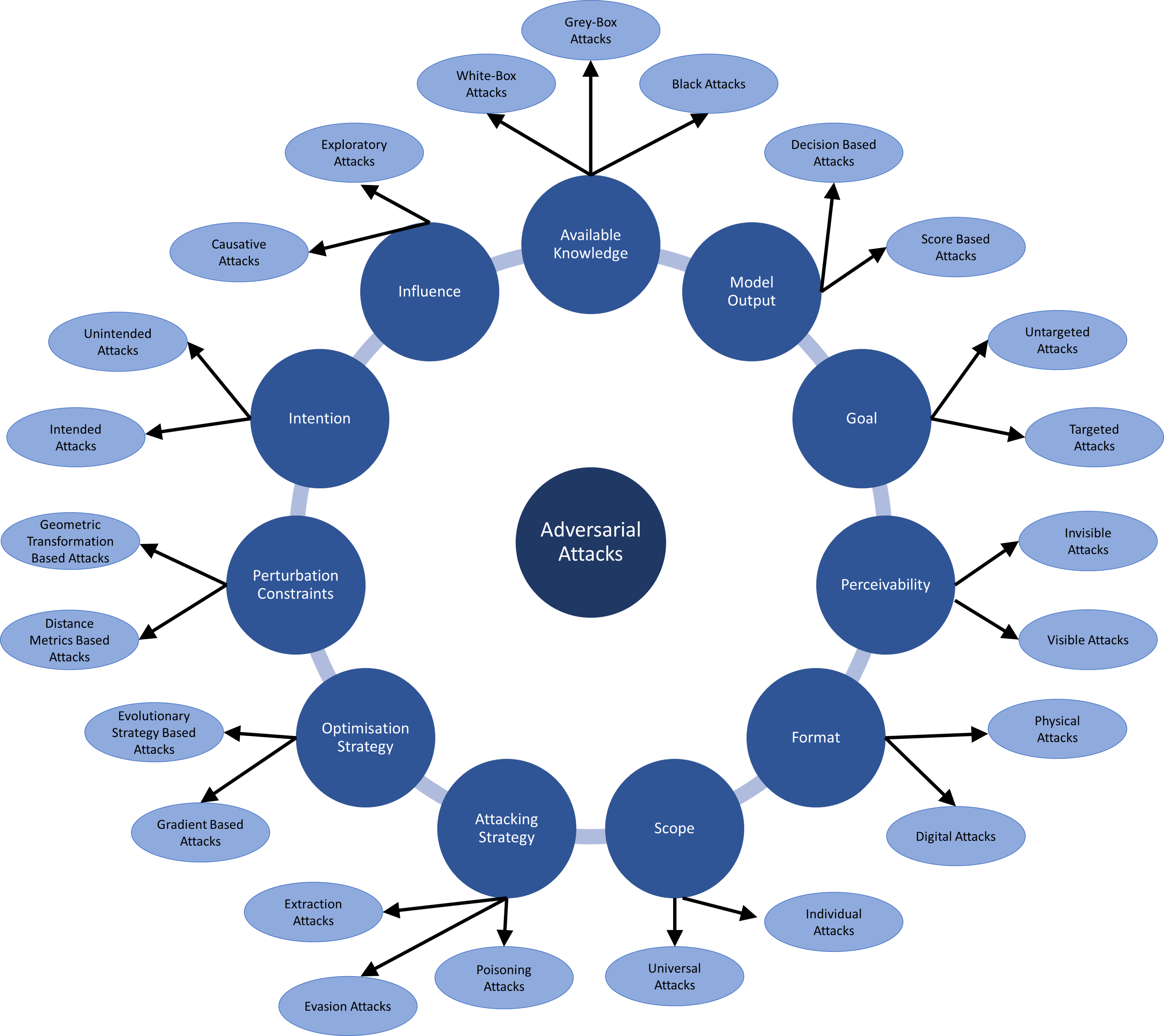}
    \caption{Classification of adversarial attacks from different perspectives.}
    \label{class}
\end{figure*}

There exist diverse types of adversarial attacks; in order to classify these several types, we employ classification of adversarial attacks from eleven different perspectives as shown in Figure \ref{class}.
An elaborate discussion on the different classifications follows,

    \subsection{Based on available knowledge to adversarial attacks}

        \begin{figure}[!t]
            \centering
            \includegraphics[width=\linewidth]{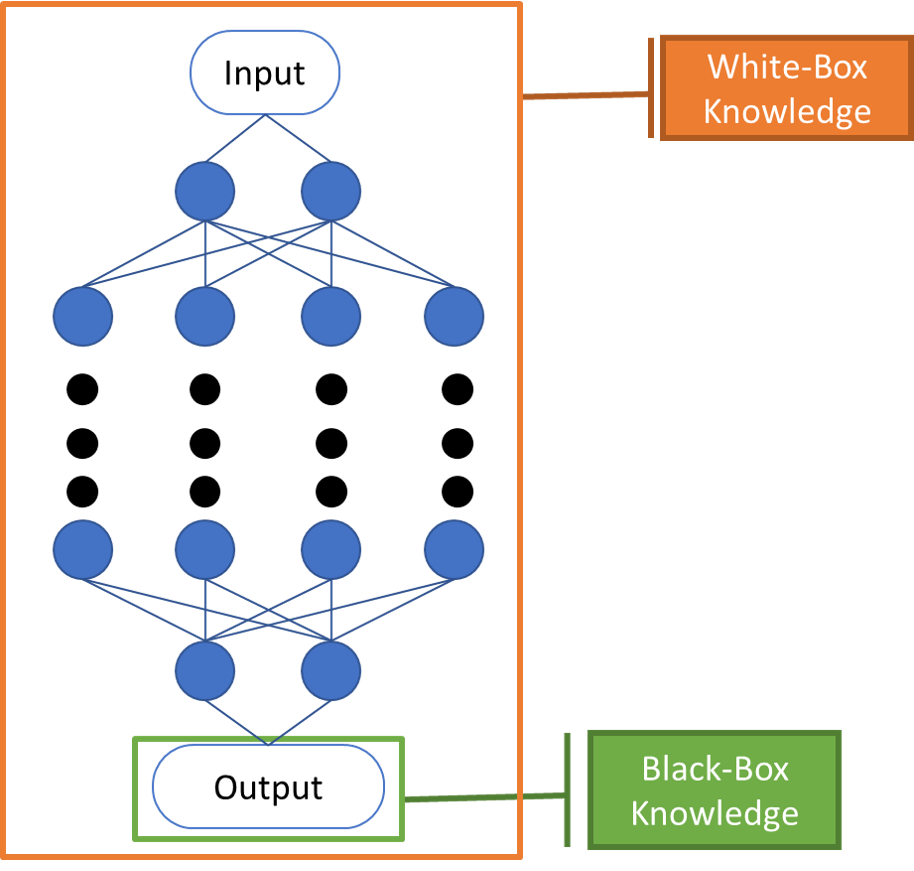}
            \caption{Knowledge of Adversarial Attack}
            \label{know}
        \end{figure}

        \begin{description}[leftmargin=*, font=$\bullet$\scshape\bfseries]

        \item[White-Box Attack (Full Knowledge):]
        In this kind of adversarial attack, the adversary has all the information about the model used, i.e., its architecture, all the individual layer parameters, and training dataset as shown in Figure \ref{know}.

        This case scenario is helpful to understand the effect of adversarial perturbations caused by the internal parameters.
        However, we have little to no knowledge of the model in the real-case scenario, making this kind of attack useful only for academic research purposes to understand the model better.
        This setting allows one to perform a worst-case evaluation of the security of learning algorithms, providing empirical upper bounds on the performance degradation that may be incurred by the system under attack.
        ~

        \item[Grey-Box Attack (Limited Knowledge):]
        In this kind of adversarial attack, the adversary has some but not all knowledge of the model, i.e., either its architecture and/or some layer parameters.

        This case scenario helps understand the relationship between the feature extractor (usually convolution network) and the classifier (usually fully connected network) parts.
        This understanding then can be used in segregating the feature extraction and classification, which is helpful for transfer learning.

        However, in most scenarios, the grey-box attack is proposed to be more potent and transferable than the white-box attack.
        Similar to white-box attacks, we have little use of these attacks in the real-case scenario, thus restricting them for academic and research purposes.

        ~

        \item[Black-Box Attack (Minimal Knowledge):]
        In this kind of adversarial attack, the adversary has no information about the model, i.e., the model only knows the output of the neural network $g(x)$ and does not know any parameters of any layer of the neural network, i.e., $f^{(i)}$ as described in Equation \ref{eqnn} and also shown in Figure \ref{know}.
        A further stronger branch of this, a no-model attack where we do not have any information from the model, input and output.

        This is the most vital kind of attack in this classification as little as possible information and is scalable to the real world, making it highly susceptible to ethical issues.
        Due to their possible scalability, some attacks can be misused in applications.
        However, most of the attacks are slower than the white-box and grey-box attacks, limiting their scalability.

        \end{description}

    \subsection{Based on model output for adversarial attacks}

        \begin{figure}[!t]
                \centering
                \includegraphics[width=\linewidth]{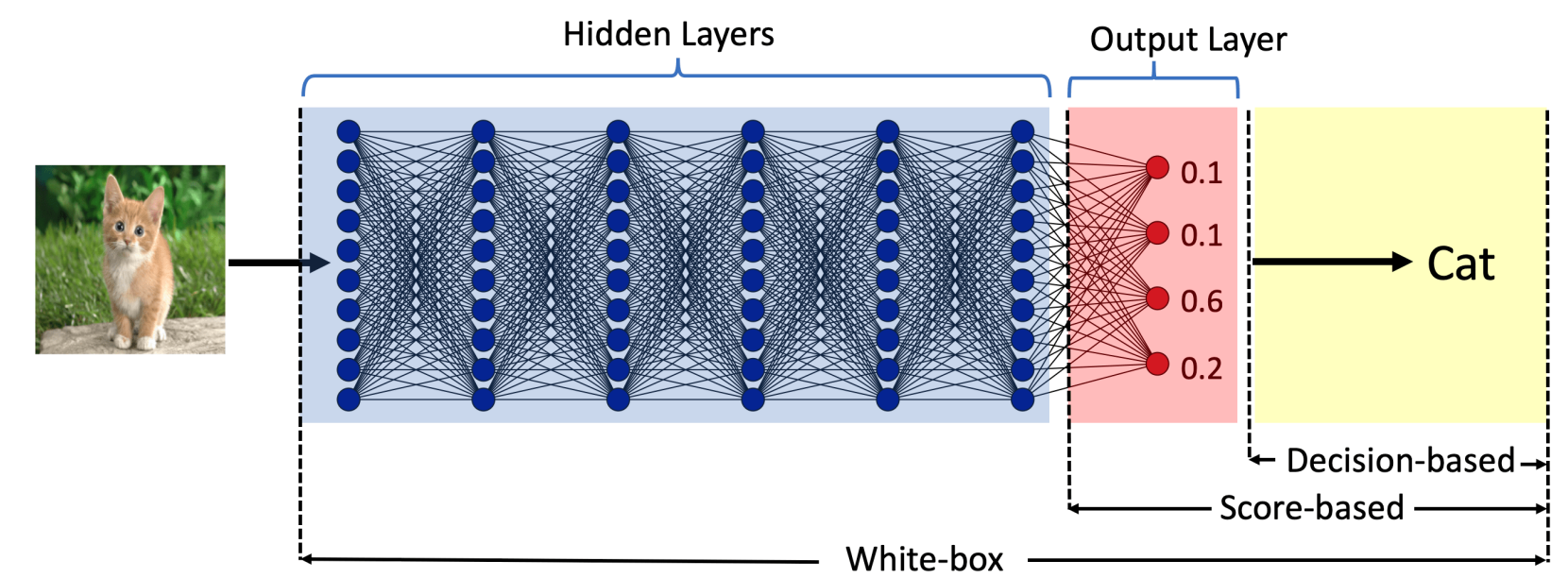}
                \caption{Difference between Decision-based, Score-based and White-box Attacks \cite{chen2020hopskipjumpattack}}
                \label{knockoff}
            \end{figure}

        \begin{description}[leftmargin=*, font=$\bullet$\scshape\bfseries]

        \item[Score based Attack:]
        In this kind of adversarial attack, the adversary knows the probabilities or soft-label for every class of the model, i.e., $g(x)$.

        ~

        \item[Decision-based Attack:]
        In this kind of adversarial attack, the adversary knows only the decision of the model, i.e., the predicted class or $\operatorname{argmax}[g(x)]$

        \end{description}

    \subsection{Based on the goal of adversarial attacks}

        \begin{description}[leftmargin=*, font=$\bullet$\scshape\bfseries]

        \item[Untargetted (Error-generic) Attack:]
        In this kind of adversarial attack, the adversary's goal is to misguide the neural network model to make a network classify any incorrect class.
        Making use of the definition of adversarial samples optimisation for the targetted attacks, can be formally defined as:
        \begin{equation*}
        \begin{aligned}
        & \text{minimize}
        & & g(x+\epsilon_{x})_C
        & \text{subject to}
        & & \epsilon_{x}
        \label{adv_eqn}
        \end{aligned}
        \end{equation*}
        where $g()_C$ is the soft-label for the correct class.
        This kind of attack is naive with the only goal for the misclassification and thus provide little information on the reasoning for the misclassification.
        At the same time, they highlight the issue of lousy representation space and the highly complex (non-smooth) decision boundary of the neural networks in higher dimensions.

        ~

        \item[Targeted (Error-Specific) Attack:]
        In this kind of adversarial attack, the adversary's goal is to misguide the model to classify a particular class other than the actual class.
        Making use of the definition of adversarial samples optimisation for the targetted attacks, can be formally defined as:
        \begin{equation*}
        \begin{aligned}
        & \text{maximize}
        & & g(x+\epsilon_{x})_T
        & \text{subject to}
        & & \epsilon_{x}
        \label{adv_eqn}
        \end{aligned}
        \end{equation*}
        where $g()_T$ is the soft-label for the target class.
        This kind of attack is harder than the untargetted attack as it can focus the attack on one particular class and help to understand the primary pattern for the class.
        However, this attack has a lower success rate than the untargeted attack making it hard to generalise for the dataset.

        \end{description}

    \begin{figure}[!t]
        \centering
        \includegraphics[width=\linewidth]{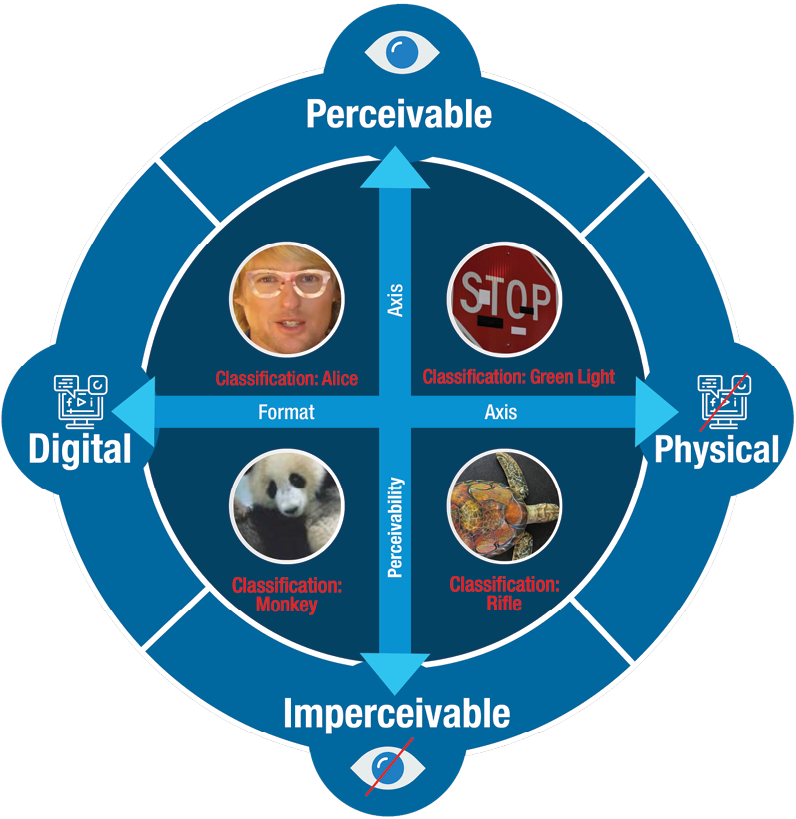}
        \caption{Difference between digital, physical, perceivable and non-perceivable attacks \cite{comiter2019attacking}}
        \label{axis}
    \end{figure}

    \subsection{Based on perceivability of adversarial samples}

        \begin{figure}[!t]
            \centering
            \includegraphics[width=\linewidth]{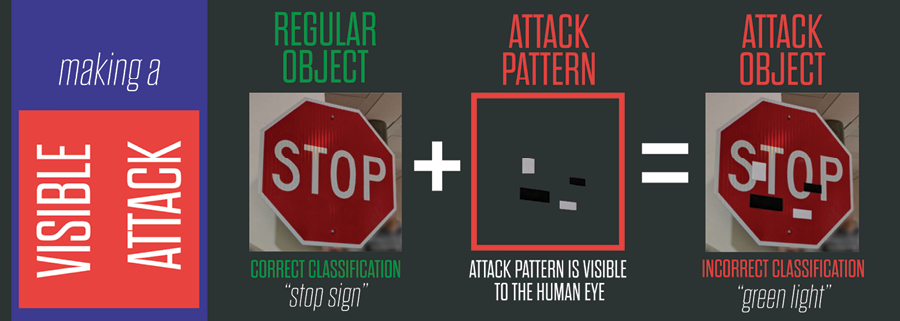} \\
            \includegraphics[width=\linewidth]{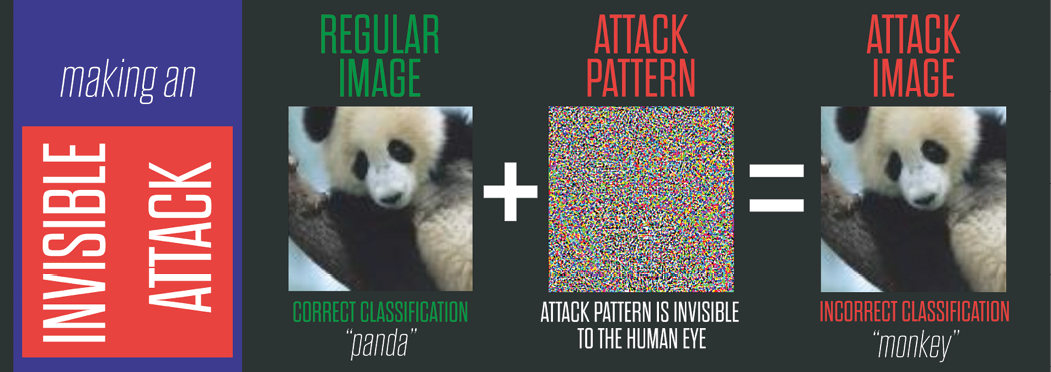}
            \caption{Example of Perceivable and Non-Perceivable Attacks \cite{comiter2019attacking}}
            \label{visible_invisible}
        \end{figure}

        \begin{description}[leftmargin=*, font=$\bullet$\scshape\bfseries]

        \item[Perceivable (Visible) Attack:]
        In this kind of adversarial attack, the perturbation is visible to the naked human eye or other senses and it usually involving adding an adversarial patch, deformation of input, or altering the input as shown in Figures \ref{axis}, and \ref{visible_invisible}.

        This kind of attack helps in understanding the difference between human learning and machine learning, where we humans tend to focus on the object of interest, machine learning attends to the broken pattern, which results in misclassification.
        The use of visible attacks also restricts real-world deployment as they can be noticed right away and can be rectified.

        ~

        \item[Non-Perceivable (Invisible) Attack:]
        In this kind of adversarial attack, the perturbation is invisible to the human senses.
        This attack usually involves the addition of digital dust on the input, which is unperceivable to the human senses as shown in Figures \ref{axis}, and \ref{visible_invisible}.

        This kind of attack helps in understanding the brittle nature of the trained model and helps us to understand the decision boundary of the model.
        At the same time, they are particularly pernicious from a security standpoint as they cannot be observed, and the model having human observation cannot be alerted by the manipulation of the input.

        \end{description}

    \subsection{Based on the format of adversarial samples}

        \begin{figure}[!t]
            \centering
            \includegraphics[width=\linewidth]{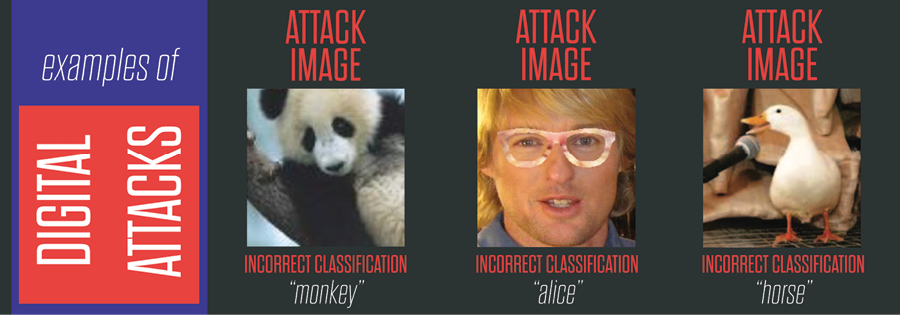} \\
            \includegraphics[width=\linewidth]{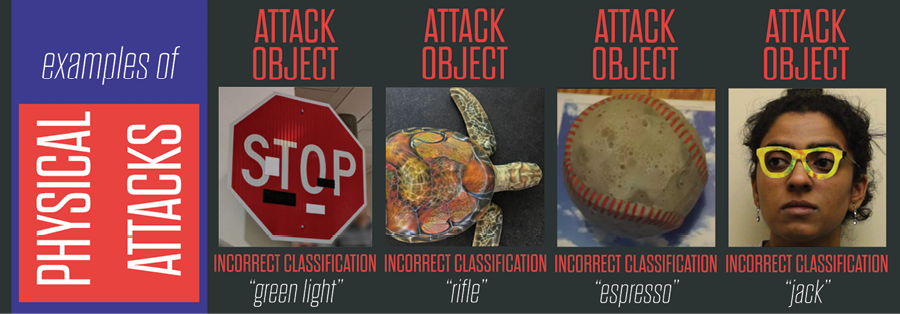}
            \caption{Example of Digital and Physical Attacks \cite{comiter2019attacking}}
            \label{digital_physical}
        \end{figure}

        \begin{description}[leftmargin=*, font=$\bullet$\scshape\bfseries]

        \item[Digital Attack:]
        In this kind of adversarial attack, the adversary targets the digital asset such as images, videos, or other files as shown in Figures \ref{axis}, and \ref{digital_physical}.
        Due to the target being a digital asset, it is a common form of attack as the adversary has expanded selection of the format and a lower difficulty of crafting adversarial perturbations.

        This kind of attack is mainly used for academic and research purposes to understand the vulnerability of neural networks.
        However, they pose an imminent threat to the models deployed as cyber hacking can lead their digital assets vulnerable to these attacks.

        ~

        \item[Physical Attack:]
        In this kind of adversarial attack, the adversary has the target in the physical world which is attacked.
        Some examples include stop signs and lane markings as shown in Figures \ref{axis}, and \ref{digital_physical}.
        Some of the physical attacks may require bigger and coarser patterns, as processing them requires digitising the physical object with a sensor.
        This digitising process may destroy finer details.

        This kind of attack is limited due to the requirement of physical objects and coarser harder attack perturbations.
        At the same time, they pose a severe security threat to a multitude of real-world applications.

        \end{description}

    \subsection{Based on the scope of adversarial attacks}

        \begin{description}[leftmargin=*, font=$\bullet$\scshape\bfseries]

        \item[Individual Attack:]
        In this kind of adversarial attack, the scope of the adversarial attack is to find a perturbation that added to one sample can misclassify the original sample, i.e. $\epsilon_{x}$ differs with the different input sample.
        In this kind of attack, optimisation is done for each input sample.

        This kind of attack helps in understanding the breakage of pattern and texture in the images. However, at the same time, due to each input sample being different, it is hard to generalise and understand the perturbation.
        Thus, it only provides us with information on robustness and leaves us with little understanding of the existence of adversarial samples.

        ~

        \item[Universal Attack:]
        In this kind of adversarial attack, the goal of the adversarial attack is to find a perturbation that added to original samples can misclassify most of the original samples, i.e., $\epsilon = \epsilon_{x_1} = \epsilon_{x_2}$ where $x_1$ and $x_2$ are different input samples and $\epsilon$ is the universal perturbation.
        In this kind of attack, optimisation is done for the entire dataset.

        This kind of attack helps in understanding the global vulnerability of the network as it finds the universal perturbation, thus finding the vulnerability in accessing the input.
        It also helps in understanding the transferability of adversarial samples.

        \end{description}

    \begin{figure}[!t]
        \centering
        \includegraphics[width=\linewidth]{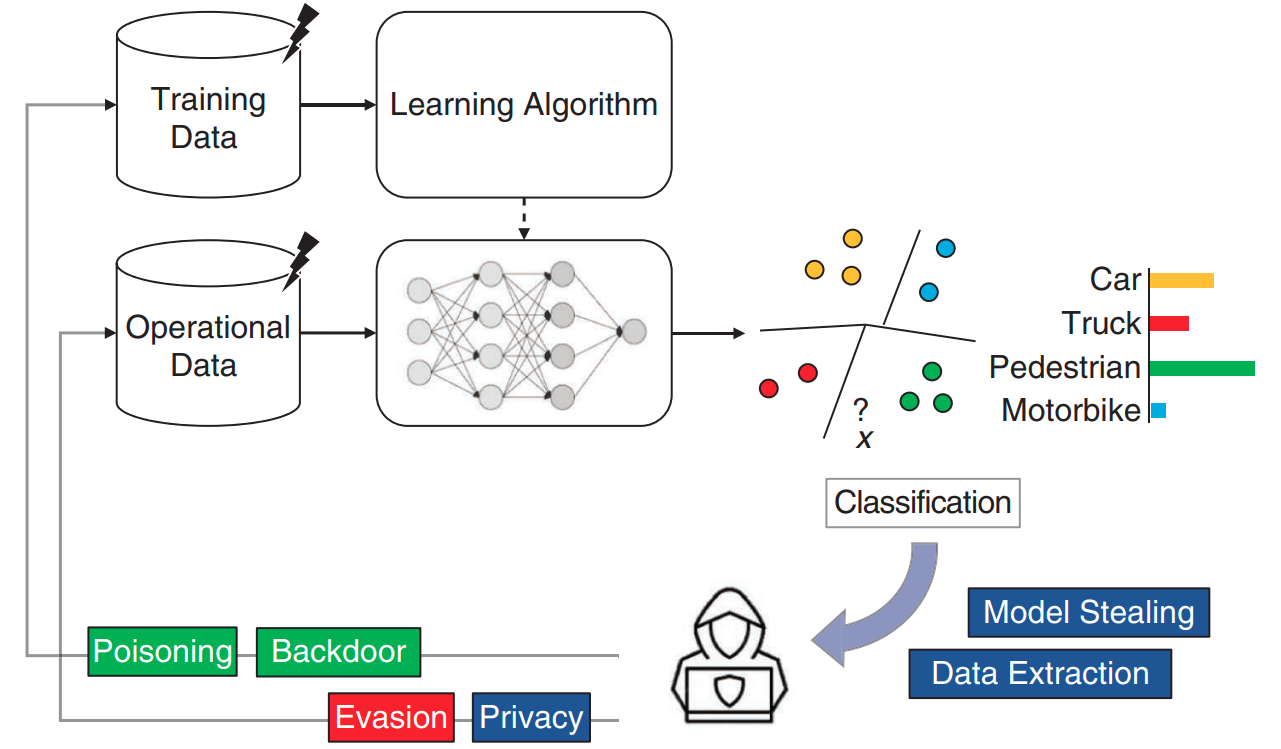}
        \caption{Illustration of poisoning, evasion and model extraction attacks \cite{lin2021adversarial}}
        \label{all}
    \end{figure}

    \subsection{Based on attacking strategy used by adversarial attacks}

        \begin{figure}[!t]
            \centering
            \includegraphics[width=\linewidth]{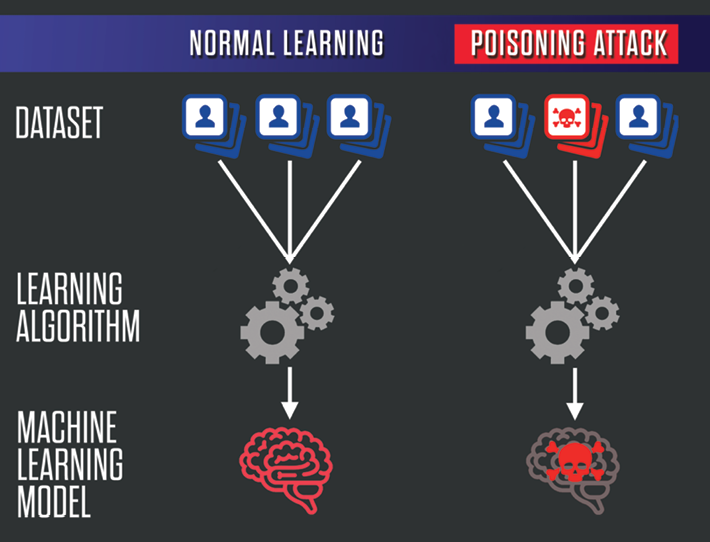}
            \caption{Illustration of data poisioning attack \cite{comiter2019attacking}}
            \label{poison}
        \end{figure}

        \begin{figure}[!t]
            \centering
            \includegraphics[width=\linewidth]{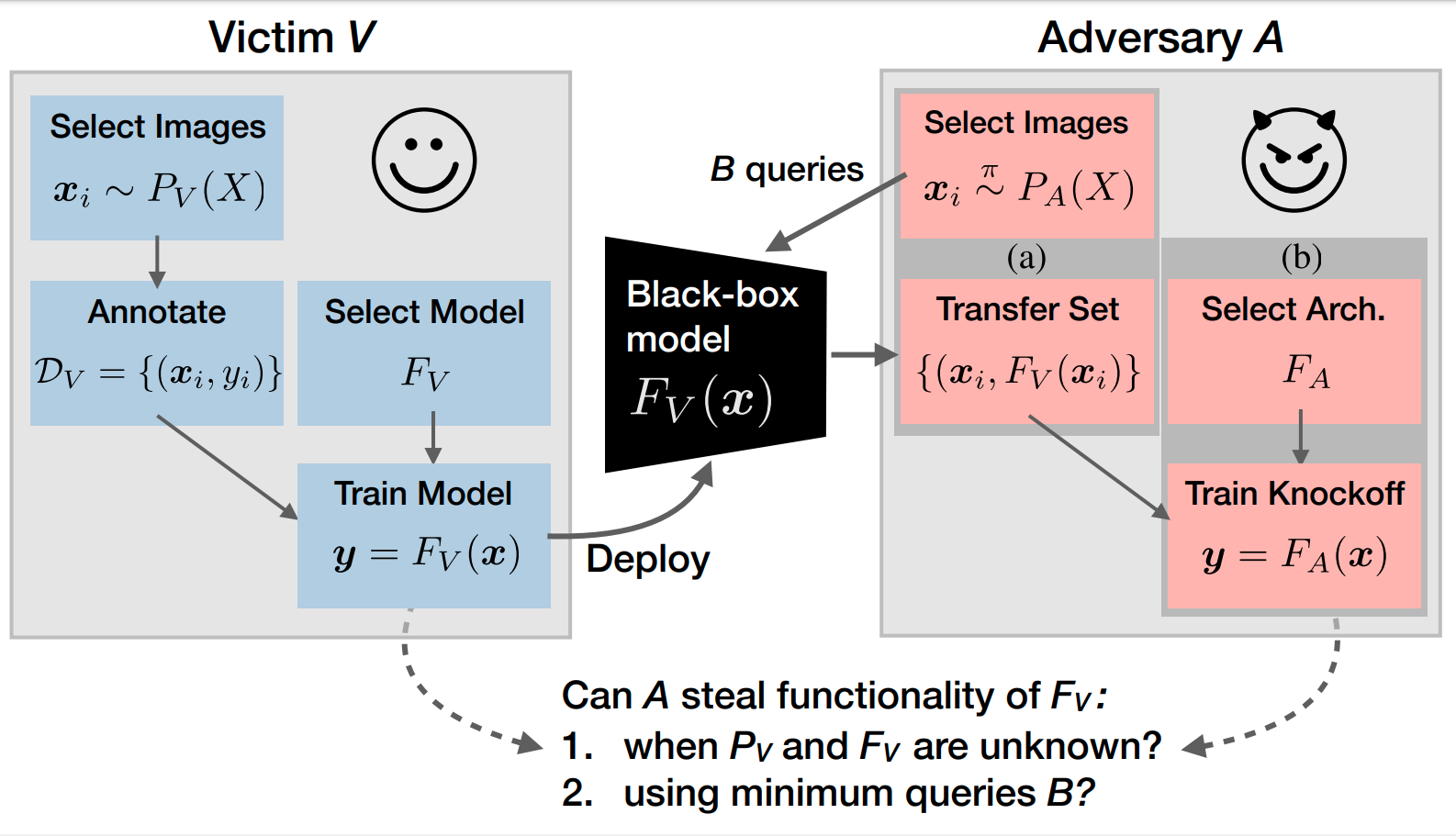}
            \caption{Problem Statement for the extraction attacks as proposed by \cite{orekondy2019knockoff}}
            \label{extract}
        \end{figure}

        \begin{description}[leftmargin=*, font=$\bullet$\scshape\bfseries]

        \item[Data Poisoning Attack:]
        In this kind of adversarial attack, the adversary usually conspires when the model is not trained, and the objective is to inject perturbations in the training data so that the model will misclassify the non-perturbed samples at the testing time as shown in Figures \ref{all}, and \ref{poison}.
        It is also known as contaminating attack.
        It relies on the compromise of the integrity of the training data and optionally availability of the model.

        Backdoor attacks are another branch of data-poisoning attacks in which adversarial attacks fool models by imprinting a texture or pattern referred to as triggers in a few training samples and changing their labels during training.
        During the inference, the misclassification is caused by injecting the trigger in the input sample.
        In a related branch of clean-label poisoning attacks, no control over the labelling process is handled to the adversarial attacks.

        Some examples include Adversarial Embedding Attack \cite{shokri2020bypassing},  Backdoor Poisioning Attack \cite{gu2017badnets} Clean Label Backdoor Attack \cite{gu2017badnets}, Bullseye Polytope Attack \cite{aghakhani2020bullseye}, Feature Collision Attack \cite{shafahi2018poison}.

        This kind of attack is effective only when the model is exposed while training, however in most cases, the trained model is deployed, making this strategy less useful in creating real-world attacks.
        However, this also helps in understanding the bias in training caused by the data and further helps us understand the decision boundary and the representation space of the model.
        Moreover, this kind of attack is effective against continual learning models and reinforcement learning models where the model interact and learn based on the outcome.

        An example case study in the Twitter chatbot `Tay' launched by Microsoft in $2016$ \cite{mathur2016intelligence}.
        The idea was to engage in conversation with users and learn playful conversations.
        However, the internet troll users started to feed profane, and offensive tweets, which resulted in the more offensive tweets by the bot \cite{ohlheiser2016trolls}.

        ~

        \item[Evasive Attack:]
        In this kind of adversarial attack, the adversary usually conspires when the model is trained, and the objective is to find perturbations for the testing data so that the neural network model will misclassify them, as shown in Figure \ref{all}.
        It is a trial and error kind of attack, and the adversary does not know which perturbations can make the model misclassify.
        It relies on the compromise of the integrity of the testing data.
        Most of the proposed attacks belong to this category.

        ~

        \item[Extraction Attack:]
        In this kind of adversarial attack, the adversary usually conspires when the model has been deployed, and the objective is to retrieve the neural network model parameters as shown in Figure \ref{extract}.
        It relies on the compromise of the confidentiality of the model as the private learned featured are recovered.

        A related branch deals in extracting the functionality of the model as shown in Figure \ref{knockoff} instead of the actual model parameters. Some of the examples include Copycat CNN \cite{correia2018copycat}, Knock-off Nets \cite{orekondy2019knockoff}, and Functionally Equivalent Extraction Attack \cite{jagielski2020high}.

        ~

        \item[Model Poisoning Attacks:]
        In this kind of adversarial attack, the adversary simply replaces the legitimate model with the adversarial one.
        This is the traditional cyber-attack in which the model file (usually weights) is replaced with another one.

        \end{description}

    \subsection{Based on optimisation strategy used by adversarial attacks}

        \begin{description}[leftmargin=*, font=$\bullet$\scshape\bfseries]

        \item[Gradient Based Attacks:]
        In this kind of adversarial attack, the adversary uses gradients and/or backpropagation to compute the perturbation required for the original sample to misclassify.
        Some of the examples include, Fast Gradient Sign Method \cite{goodfellow2014explaining}, and Projected Gradient Descent Attack \cite{madry2018towards}.

        ~

        \item[Evolutionary Strategy Based Attacks:]
        In this kind of adversarial attack, adversery use evolutionary-strategy based optimisation like Differential Evolution (DE) \cite{storn1997differential}, and Covariance Matrix Adaptation Evolution Strategy (CMA-ES) \cite{hansen2003reducing} to search for the adversarial perturbation $\epsilon_{x}$.
        This is done for the black-box attacks, where model information is not present, and optimisation cannot rely on backpropagating gradients.
        Some of the examples include One-Pixel Attack \cite{su2019one} and Threshold Attack \cite{vargas2019robustness}.

        \end{description}

    \subsection{Based on perturbation constraints on adversarial samples}

        \begin{table}[!t]
            \centering
            \caption{Distance metrics based attacks used in literature compiled by Kotyan et al. \cite{vargas2019robustness}.}
            \label{related}
            \resizebox{\columnwidth}{!}{
            \begin{tabular}{l|llll}
            \toprule
            \textbf{Literature} & \textbf{$L_0$} & \textbf{$L_1$} & \textbf{$L_2$} & \textbf{$L_\infty$} \\
            \midrule

            Chen et al. (2020) \cite{chen2020hopskipjumpattack}      &  &  & \checkmark & \checkmark \\
            Croce and Hein (2020) \cite{croce2020reliable}           &  &  & \checkmark & \checkmark \\
            Ghiasi et al. (2020) \cite{ghiasi2020breaking}           &  &  & \checkmark & \checkmark \\
            Hirano and Takemoto (2020) \cite{hirano2020simple}       &  &  & \checkmark &  \\

            \midrule

            Cohen et al. (2019) \cite{cohen2019certified}            &  &  & \checkmark &  \\
            Kotyan et al. (2019) \cite{vargas2019robustness}         &  &  &  & \checkmark \\
            Su et al. (2019) \cite{su2019one}                        & \checkmark &  &  & \\
            Tan and Shokri (2019) \cite{tan2019bypassing}            & \checkmark &  &  & \\
            Wang et al. (2019) \cite{wang2019neural}                 &  &  & \checkmark &  \\
            Wong et al. (2019) \cite{wong2019wasserstein}            &  &  &  & \checkmark \\
            Zhang et al. (2019a) \cite{zhang2019towards}             &  &  &  & \checkmark \\
            Zhang et al. (2019b) \cite{zhang2019theoretically}       &  &  &  & \checkmark \\

            \midrule

            Brendel et al. (2018) \cite{brendel2018decision}         &  &  & \checkmark & \\
            Buckman et al. (2018) \cite{buckman2018thermometer}      &  &  &  & \checkmark \\
            Gowal et al. (2018) \cite{gowal2018effectiveness}        &  &  &  & \checkmark \\
            Grosseet al. (2018) \cite{grosse2018limitations}         &  &  & \checkmark &  \\
            Guo et al. (2018) \cite{guo2018countering}               &  &  & \checkmark & \checkmark \\
            Madry et al. (2018) \cite{madry2018towards}              &  &  & \checkmark & \checkmark \\
            Singh et al. (2018) \cite{singh2018fast}                 &  &  &  & \checkmark \\
            Song et al. (2018) \cite{song2018pixeldefend}            &  &  &  & \checkmark \\
            Tramer et al. (2018) \cite{tramer2018ensemble}           &  &  &  & \checkmark \\

            \midrule

            Arpit et al. (2017) \cite{arpit2017closer}               &  &  &  & \checkmark \\
            Carlini and Wagner (2017) \cite{carlini2017towards}      & \checkmark  &  & \checkmark  & \checkmark \\
            Chen et al. (2017a) \cite{chen2017ead}                   &  & \checkmark & \checkmark & \checkmark \\
            Chen et al. (2017b) \cite{chen2017zoo}                   &  &  & \checkmark &  \\
            Das et al. (2017) \cite{das2017keeping}                  &  &  & \checkmark & \checkmark \\
            Gu et al. (2017) \cite{gu2017badnets}                    & \checkmark &  & \checkmark & \\
            Jang et al. (2017) \cite{jang2017objective}              &  &  &  & \checkmark \\
            Moosavi et al. (2017) \cite{moosavi2017universal}        &  &  & \checkmark & \checkmark \\
            Xu et al. (2017) \cite{xu2017feature}                    & \checkmark &  & \checkmark & \checkmark \\

            \midrule

            Kurakin et al. (2016) \cite{kurakin2016adversarial}      &  &  &  & \checkmark \\
            Moosavi et al. (2016) \cite{moosavi2016deepfool}         &  &  & \checkmark & \checkmark \\
            Papernot et al. (2016a) \cite{papernot2016limitations}   & \checkmark &  &  & \\
            Papernot et al. (2016b) \cite{papernot2016distillation}  & \checkmark &  &  & \\

            \midrule

            Goodfellow et al. (2014) \cite{goodfellow2014explaining} &  &  &  & \checkmark \\

            \bottomrule
            \end{tabular}
            }
        \end{table}

        \begin{description}[style=sameline, leftmargin=*, font=$\bullet$\scshape\bfseries]

        \item[Distance metrics based constraint Attack:]
        In this kind of adversarial attack, the adversary constrains the perturbation search for the input by using distance from the original sample.
        Several different distance norms are used to create different types of attacks.
        Here, the perturbation $\epsilon_{x}$ is constrained such that $\Vert \epsilon_{x} \Vert_p < th$, where $p$ is the distance norm and $th$ is the threshold constraint.
        Some of the common norms used are $L_0$, $L_1$, $L_2$, and $L_\infty$, and Table \ref{related} shows examples of adversarial attacks employing different distance norms.

        ~

        \item[Geometric Transformations based constraint Attack:]
        In this kind of adversarial attack, the adversary relies on geometric transformations of the original input to misclassify the model.
        Spatial Transformations like affine and rotations can misclassify the network even though the network is trained using those as augmentations in the training dataset \cite{engstrom2019exploring}.
        Color-channel permutation also affects the model's performance and can be used as an adversarial attack \cite{kantipudi2020color}.

        \end{description}

    \subsection{Based on intention of adversarial attacks}

        \begin{description}[leftmargin=*, font=$\bullet$\scshape\bfseries]

        \item[Intented Attack:]
        In this kind of adversarial attack, the adversary has the malicious intention of misclassifying the model based on perturbations.

        ~

        \item[Unintented Attack:]
        The model misclassifies this kind of adversarial attack due to the various unintended failures in training or inference.
        An example includes reward hacking in which reinforcement learning systems act weirdly because of the discrepancies between the specified reward received by the model and the true intended reward \cite{kumar2019failure}.

        \end{description}

    \subsection{Based on adversarial attacker's influence}

        \begin{description}[leftmargin=*, font=$\bullet$\scshape\bfseries]

        \item[Causative Attack:]
        In this kind of adversarial attack, the adversary can manipulate both training and testing data.

        ~

        \item[Exploratory Attack:]
        In this kind of adversarial attack, the adversary can only manipulate the testing data.

        \end{description}

\section{Understanding adversarial attacks}

    It is hypothesised in \cite{goodfellow2014explaining} that neural networks' linearity is one of the principal reasons for failure against an adversary and non-linear neural networks are thus, more robust compared to linear networks \cite{guo2018sparse}.
    Based on this understanding, it is proposed in \cite{buckman2018thermometer} discretise the input feature space, which may lead to breaking this linearity.

    A geometric perspective is analysed in \cite{moosavi2018robustness}, where it is shown that adversarial samples lie in shared subspace, along which the decision boundary of a classifier is positively curved.
    Further, in \cite{fawzi2018empirical}, a relationship between sensitivity to additive perturbations of the inputs and the curvature of the decision boundary of deep networks is shown.

    Another aspect of robustness is discussed in \cite{madry2018towards}, where authors suggest that the capacity of the neural networks' architecture is relevant to the robustness.
    However, research shows that the input feature space itself is vast, which provide opportunities to the adversaries \cite{xu2017feature}.
    It was also observed that the classifiers are not familiarised with the adversarial input feature space as adversarial samples have much lower probability densities under the image distribution \cite{song2018pixeldefend}.

    Intuitively; thus, in \cite{das2017keeping} authors recommended discarding some of the information unnoticeable to humans in input feature space by compressing as adversarial noises are often indiscernible by the human eye.
    The bounds for the robustness using this input feature space is also studied in \cite{fawzi2018adversarial}.
    Further, the existence of different internal representations learned by neural networks for an adversarial sample compared to a benign sample is shown in \cite{sabour2015adversarial}.

    It is also stated in \cite{ilyas2019adversarial} that the adversarial vulnerability is a significant consequence of the dominant supervised learning paradigm and a classifier's sensitivity to well-generalising features in the known input distribution.
    Also, research by \cite{tao2018attacks} argues that adversarial attacks are entangled with the interpretability of neural networks as results on adversarial samples can hardly be explained.

    Another investigation proposes the conflicting saliency added by adversarial samples as the reason for misclassification \cite{vargas2019understanding}.
    It was shown in \cite{kotyan2021deep} that perturbation causes a shift in attention of the neural network, which is a probable cause for the misclassification.
    While the white-box gradient-based attack consistently scattered the attention from the object of interest, the black-box attack was shown to either bring the model's attention to the perturbation or disrupt the attention around the perturbation.

\section{Tackling adversarial attacks with adversarial defences}

    Many defensive systems and detection systems have also been proposed to mitigate some of the problems.
    Some approaches rely on detecting adversarial samples to mitigate the adverse effects of adversarial algorithms, while some approaches rely on defensive algorithms.
    However, there are still no current solutions or promising ones which can negate the adversarial attacks \textit{consistently}.
    Regarding defensive systems, there are many variations of defenses \cite{dziugaite2016study,hazan2016perturbations,das2017keeping,guo2018countering,song2018pixeldefend,xu2017feature,ma2018characterizing,buckman2018thermometer}
    which are carefully analysed in \cite{athalye2018obfuscated,uesato2018adversarial} and many of their shortcomings are documented.

    Defensive distillation \cite{papernot2016distillation}, a defence was proposed, in which a smaller neural network squeezes the content learned by the original one was proposed as a defence.
    However, it was shown not to be robust enough in \cite{carlini2017towards}.
    Adversarial training was also proposed, in which adversarial samples are used to augment the training dataset \cite{goodfellow2014explaining}, \cite{madry2018towards}.
    Augmentation of the dataset is done so that the neural network should classify the adversarial samples, thus increasing their robustness.
    Although adversarial training can increase the robustness slightly, the resulting neural network is still vulnerable to attacks \cite{tramer2018ensemble}.

    Regarding detection systems, a study from \cite{grosse2017statistical} demonstrated that indeed some adversarial samples have different statistical properties which could be exploited for detection.
    The authors in \cite{xu2017feature} proposed to compare the prediction of a classifier with the prediction of the same input, but `squeezed'.
    This technique allowed classifiers to detect adversarial samples with small perturbations.
    Many detection systems fail when adversarial samples deviate from test conditions \cite{carlini2017adversarial,carlini2017magnet, carlini2019evaluating}.
    Thus, the clear benefits of detection systems remain inconclusive.

    It was shown in \cite{vargas2019understanding} that changes in pixels of an image propagate and expand throughout the layers to either disappear or cause significant changes in the classification.
    It was also shown that perturbation in nearby pixels of successful one-pixel attack has high attack accuracy.
    This suggests that changes in a pixel may increase or decrease the influence of a receptive field (small group of nearby pixels).
    This is a direct relationship of the convolution, which is a linear operation.

    In the adversarial setting, the analysis of the spatial distribution of saliency proves helpful to interpret why changing some pixels \cite{su2019one} in the network corresponds to misclassification.
    It was hypothesised in \cite{vargas2019understanding} that the existence of adversarial samples is due to conflicting saliency, which causes enough disturbance in the neural network forcing it to misclassify.
    Hence, adversarial samples are not naively fooling neural networks but diverting their attention towards another part of the image.

\section{Conclusion}

In this survey, we analysed the importance and the significance of adversarial attacks as a real-world threat to applications in different domains.
In order to tackle the imminent threat, it is required to understand the theoretical and empirical limitations and effects of adversarial attacks.
In order to facilitate the understanding, we provide an extensive classification from various perspectives and discuss their implications.
Further, we also provide a brief survey of literature on understanding adversarial attacks and adversarial defences.

Based on our understanding, we can now define few research directions issues which are yet to be resolved.
\begin{description}[leftmargin=*, font=$\bullet$\scshape\bfseries]

    \item[Adative adversarial attacks and defences:]
    Current adversarial attacks and defences are static in nature and do not adapt to changing scenarios.
    This limits the use of both adversarial attacks and mitigating their adverse effects by adversarial defences.

    \item[Exploring black-box nature:]
    Current state-of-the-art models are black-box in nature which makes auditing their output difficult.
    This limits our understanding of the existence of adversarial examples as well as our understanding to identify if an adversarial attack has compromised the model.

    \item[Learning resilient features:]
    The current state of the art models is brittle in nature as it is easy to disrupt the learned features.
    Supervised end-to-end learning needs to be replaced with different learning techniques which are more robust and less brittle in nature.

    \item[Proactive measures:]
    Current adversarial defences are reactive in nature where they defend against an existing adversarial attack.
    However, a lack of proactive framework remains a challenge to handle the new adversarial attacks.

    \item[Theroretical limitations:]
    Current neural networks are difficult to analyse theoretically due to their complicated non-convex properties.
    Further, adversarial samples are solutions to the optimisation problem, which is non-linear and non-convex.
    Because of our limited theoretical tools, it is hard to describe the solutions to these complicated optimisation problems and further complex to create a theoretical framework for adversarial defences which can mitigate a set of adversarial samples.
\end{description}

\section{Timeframe of survey}
This article covers the adversarial attacks for neural networks and related works published between $2013$ and $2020$.

\section{Further readings}

\begin{description}[leftmargin=*, font=$\bullet$\scshape\bfseries]

\item[Article:] On evaluating adversarial robustness by Carlini et al. \cite{carlini2019evaluating}

~

\item[Article:] Wild patterns: Ten years after the rise of adversarial machine learning by Biggio et al. \cite{biggio2018wild}.

~

\item[Article:] Obfuscated gradients give a false sense of security: Circumventing defences to adversarial examples by Athalye et al. \cite{athalye2018obfuscated}.

~

\item[Technical Report:] The Malicious Use of Artificial Intelligence: Forecasting, Prevention, and Mitigation by Brundage et al. \cite{brundage2018malicious}

~

\item[Book:] Adversarial Machine Learning by Joseph et al. \cite{huang2011adversarial}.

~

\item[Resource:] CleverHans library, compiled by Tensorflow and managed by Nicolas Paper not, Ian Goodfellow and others, is an adversarial example library for ``constructing attacks, building defences, and benchmarking both'' \cite{papernot2018cleverhans}.

~

\item[Resource:] Adversarial Robustness Toolbox library, compiled by IBM, is a library for various adversarial attacks, defences, and metrics used \cite{art2018}.

~

\item[Resource:] Foolbox library compiled by Bethge Lab \cite{rauber2017foolboxnative,rauber2017foolbox}

~

\item[Competition:] Unrestricted Adversarial Examples Contest was sponsored by Google Brain, which was ``a  community-based challenge to incentivise and measure progress towards the goal of zero confident classification errors in machine learning models". \url{https://ai.googleblog.com/2018/09/introducing-unrestricted-adversarial.html}

~

\item[Article List:] Nicolas Carlini, a leading researcher in adversarial machine learning, maintains an unfiltered list of $1000+$ articles around the field of adversarial machine learning and also maintains a curated list of articles for the introduction in the field. \url{https://nicholas.carlini.com/writing/2018/adversarial-machine-learning-reading-list.html}

~

\item[Research Group:] Madry lab, led by Aleksander Mądry of Massachusetts Institute of Technology, is a leading research group in the field of adversarial machine learning with the focus on understanding the robustness of neural networks. \url{http://madry-lab.ml/}

~

\item[Research Group:] Bethge lab, led by Matthias Bethge of the University of Tübingen, is a leading research group in understanding the characteristics of neural networks. \url{http://bethgelab.org/}

~

\item[Research Group:] Trusted AI group of IBM Research AI is a leading research group focusing on instilling more trust in the decisions of the neural networks by tackling issues like fairness, robustness, explainability, transparency and accountability. \url{https://www.research.ibm.com/artificial-intelligence/trusted-ai/}

~

\end{description}

\bibliographystyle{IEEEtran}

\bibliography{adversarial_machine_learning}

\end{document}